\documentclass[10pt,twocolumn,letterpaper]{article}

\usepackage[pagenumbers]{cvpr} 

\usepackage{microtype}

\renewcommand{\paragraph}[1]{\vspace{.25em}\noindent\textbf{#1.}}

\usepackage[accsupp]{axessibility} 

\usepackage{multirow}
\usepackage{colortbl}
\usepackage{array}

\newcommand{\segw}{0.19\columnwidth}
\newcommand{\cmark}{\textcolor[RGB]{0,160,0}{\checkmark}}
\newcommand{\xmark}{\textcolor{red}{$\times$}}

\definecolor{colseg}{RGB}{218,232,252}    
\definecolor{colreg}{RGB}{253,229,216}    
\definecolor{coldann}{RGB}{232,222,246}   
\definecolor{rowours}{RGB}{226,240,227}   
\definecolor{rowgroup}{gray}{0.93}        

\usepackage{tikz}
\usepackage{pgfplots}
\usepackage{pgfplotstable}
\pgfplotsset{compat=1.18}
\usetikzlibrary{shapes.geometric, arrows.meta, positioning, fit, backgrounds, calc}

\definecolor{cvprblue}{rgb}{0.21,0.49,0.74}
\usepackage[pagebackref,breaklinks,colorlinks,allcolors=cvprblue]{hyperref}

\def\paperID{17} 
\def\confName{VISION26}
\def\confYear{2026}

\title{FryNet: Dual-Stream Adversarial Fusion for Non-Destructive Frying Oil Oxidation Assessment}

\author{
Khaled R Ahmed$^1$,
Toqi Tahamid Sarker$^1$,
Taminul Islam$^1$,
Tamany M Alanezi$^{1,2}$,
Amer AbuGhazaleh$^1$\\
$^1$Southern Illinois University Carbondale, USA \quad $^2$Qassim University, Saudi Arabia\\
{\tt\small \{khaled.ahmed, toqitahamid.sarker, taminul.islam, tamany.alanezi, aabugha\}@siu.edu}
}

\begin{document}
\maketitle
\begin{abstract}
Monitoring frying oil degradation is critical for food safety, yet
current practice relies on destructive wet-chemistry assays that
provide no spatial information and are unsuitable for real-time use.
We identify a fundamental obstacle in thermal-image-based inspection,
the camera-fingerprint shortcut, whereby models memorize
sensor-specific noise and thermal bias instead of learning oxidation
chemistry, collapsing under video-disjoint evaluation.
We propose FryNet, a dual-stream RGB-thermal framework that jointly
performs oil-region segmentation, serviceability classification, and
regression of four chemical oxidation indices (PV, p-AV, Totox,
temperature) in a single forward pass.
A ThermalMiT-B2 backbone with channel and spatial attention
extracts thermal features, while an RGB-MAE Encoder learns
chemically grounded representations via masked autoencoding and
chemical alignment.
Dual-Encoder DANN adversarially regularizes both streams against
video identity via Gradient Reversal Layers, and FiLM fusion bridges
thermal structure with RGB chemical context.
On 7{,}226 paired frames across 28 frying videos, FryNet achieves
98.97\% mIoU, 100\% classification accuracy, and 2.32 mean
regression MAE, outperforming all seven baselines.
\end{abstract}

\section{Introduction}
\label{sec:intro}

Frying oil degrades through thermal oxidation, accumulating harmful
aldehydes and polar compounds that compromise food safety once regulatory
thresholds are exceeded~\cite{lizhi2019dielectric}.
Compliance today relies on destructive wet-chemistry assays such as
titration for PV and spectrophotometry for p-AV, which take hours, consume reagents, and
yield a single scalar per sample with no spatial information about
\emph{where} degradation is occurring in the fryer.
Imaging-based non-destructive testing (NDT) can overcome all
three limitations by providing real-time, spatially resolved quality maps
without contact.
Each existing modality, however, addresses only part of the problem:
NIR spectroscopy requires costly point-probe
hardware~\cite{cayuela2025nir}, RGB systems capture only surface-color
proxies~\cite{udomkun2019cvs}, and thermal imaging has so far been
applied to oil-type classification rather than quantitative oxidation
regression~\cite{pirola2024infrared}.
No prior method fuses RGB and thermal streams for direct, dense
prediction of chemical oxidation indices across unseen oil batches.

\begin{figure}[t]
\centering
\includegraphics[width=\columnwidth]{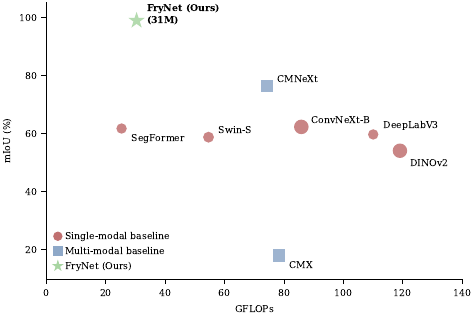}
\caption{mIoU vs.\ computational cost (GFLOPs).
         Marker size is proportional to parameter count.
         FryNet (31M params, 30.3 GFLOPs) achieves the highest mIoU
         at lower computational cost than all comparable methods.}
\label{fig:miou_efficiency}
\vspace{-10pt}
\end{figure}

Closing this gap, however, exposes a hidden confound.
In our 28-video thermal dataset, each video captures a single frying batch
and therefore carries a single quality label.
Thermal cameras simultaneously imprint device-specific signatures (sensor
noise, vignetting, thermal bias) that are constant within each video.
Because each video pairs exactly one camera with exactly one label,
a model can reach 97\% training segmentation accuracy within
4{,}000 iterations by memorizing camera fingerprints rather than
learning oxidation chemistry.
Under video-disjoint evaluation on unseen batches, five backbone
architectures (25M to 88M parameters) all collapse to 54--62\% mIoU.
This \emph{camera-fingerprint shortcut} extends beyond the thermal stream: adding an RGB stream without regularization introduces a second
shortcut channel that negates the benefit of fusion entirely.

\begin{figure*}[t]
\centering
\includegraphics[width=\textwidth]{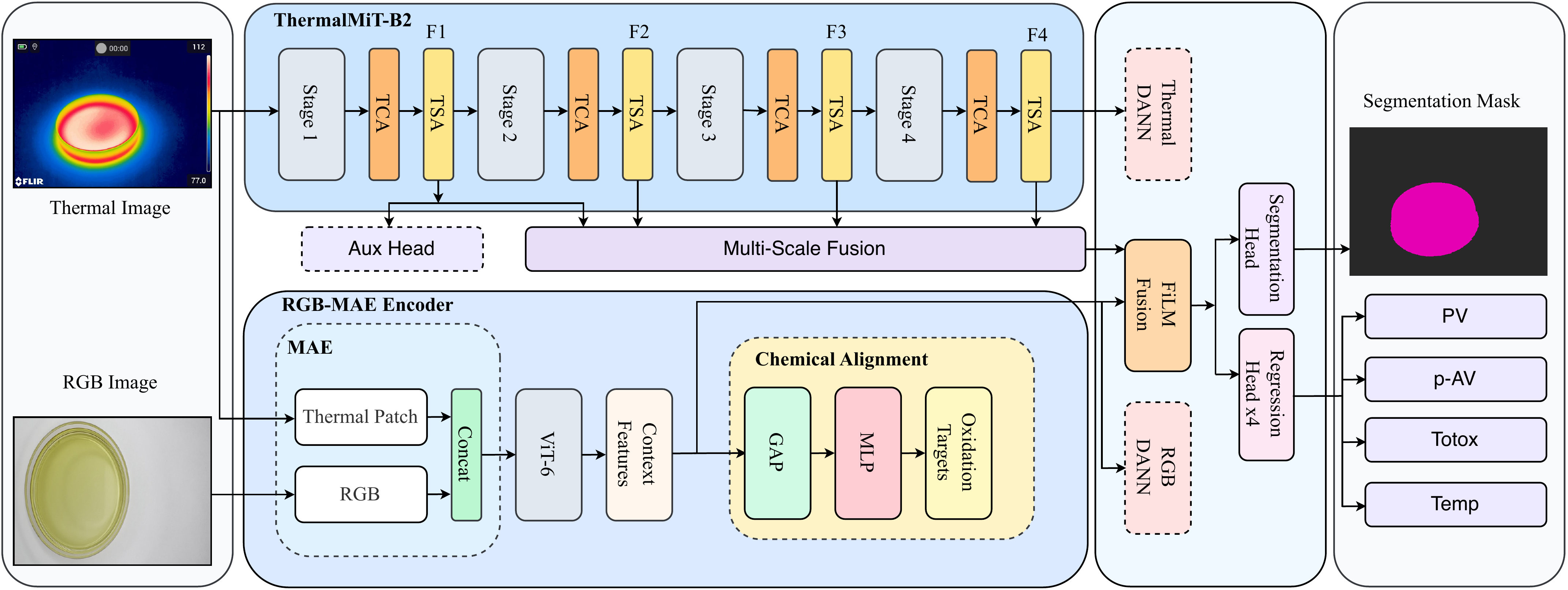}
\caption{
  \textbf{FryNet architecture.}
  The thermal stream (top) processes FLIR images through ThermalMiT-B2 with TCA/TSA attention at each stage, producing multi-scale features $F_1$--$F_4$ that are merged via multi-scale fusion.
  The RGB-MAE encoder (bottom) concatenates thermal patches with RGB inputs and learns context features via masked autoencoding with chemical alignment.
  FiLM fusion combines both streams before feeding into the segmentation head and four regression heads (PV, p-AV, Totox, Temp).
  Dual-DANN branches apply gradient reversal to both thermal and RGB features for domain-invariant learning.
  Dashed boxes indicate training-only components.
}
\label{fig:architecture}
\vspace{-10pt}
\end{figure*}

We present FryNet, a dual-stream RGB-thermal system that addresses this
shortcut while performing segmentation, classification, and regression in
a single forward pass (Fig.~\ref{fig:architecture}).
Gradient Reversal Layers~\cite{ganin2016dann} adversarially regularize
both the thermal backbone and the RGB encoder against video identity,
forcing each stream to discard camera-specific nuisances.
A cross-modal chemical alignment loss grounds
the RGB encoder in oxidation chemistry, and fused regression routing lets
all chemical targets benefit from RGB context.
FryNet achieves 98.97\% mIoU and a mean regression MAE of 2.32
(Figure~\ref{fig:miou_efficiency}), a 3.2$\times$ improvement over the best single-modal baseline.
Our main contributions are:
\begin{enumerate}
  \item We identify a camera-fingerprint shortcut that
        collapses five architectures to 54--62\% mIoU under
        video-disjoint evaluation, and show that unregularized RGB
        fusion amplifies the failure.

  \item We propose FryNet, a dual-stream multi-task
        architecture with a ThermalMiT-B2 backbone (TCA/TSA
        attention), an RGB-MAE Encoder for cross-modal representation
        learning, FiLM fusion, and Dual-Encoder DANN with a
        chemical alignment loss that suppresses shortcuts in both
        streams.

  \item We release the FryNet dataset: 7{,}226 RGB-thermal
        frame pairs across 28 videos with segmentation masks,
        serviceability labels, and four regression targets.
\end{enumerate}

\section{Related Work}
\label{sec:related}

\begin{figure}[t]
\centering
\setlength{\tabcolsep}{1pt}
\tiny
\begin{tabular}{@{}c cc@{}}
& Thermal & RGB \\[1pt]
\raisebox{0.07\textwidth}{\rotatebox[origin=c]{90}{\scriptsize\textbf{\textcolor[HTML]{A8D5A2}{Good}}}} &
\includegraphics[width=0.16\textwidth]{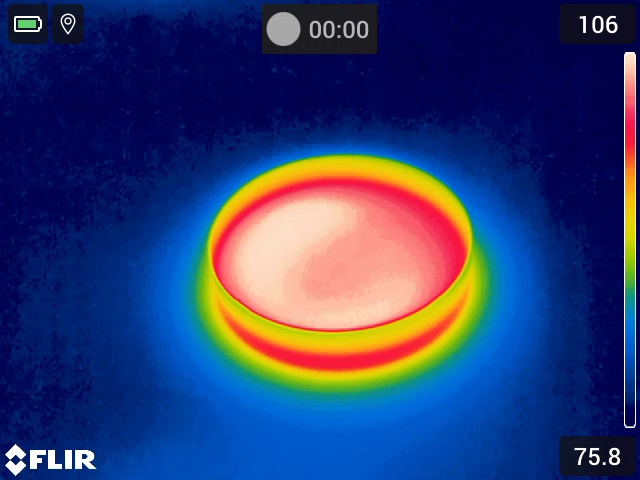} &
\includegraphics[width=0.16\textwidth]{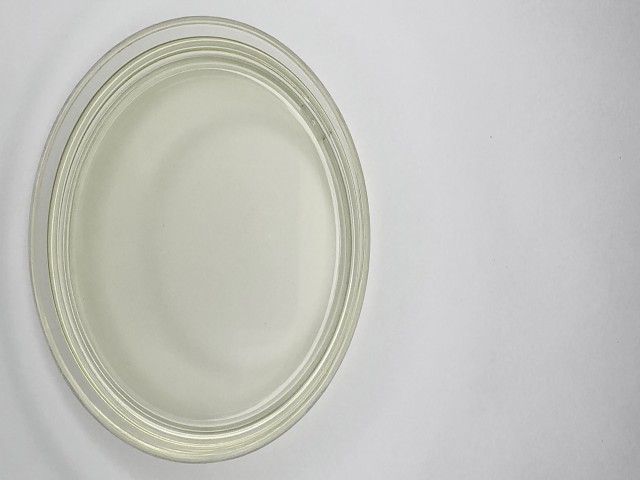} \\[-3pt]
& \multicolumn{2}{c}{%
  \colorbox{black!5}{\tiny
    PV\,=\,5.0 ~ p-AV\,=\,5.1 ~ Totox\,=\,15.0 ~ Temp\,=\,106\textdegree F}} \\[2pt]
\raisebox{0.07\textwidth}{\rotatebox[origin=c]{90}{\scriptsize\textbf{\textcolor[HTML]{C07070}{Replace}}}} &
\includegraphics[width=0.16\textwidth]{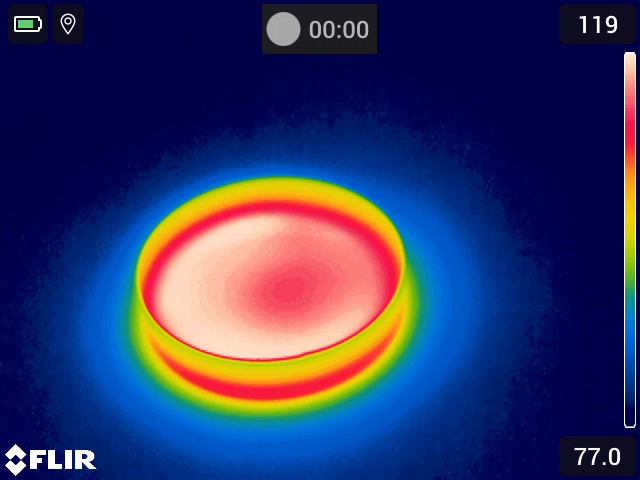} &
\includegraphics[width=0.16\textwidth]{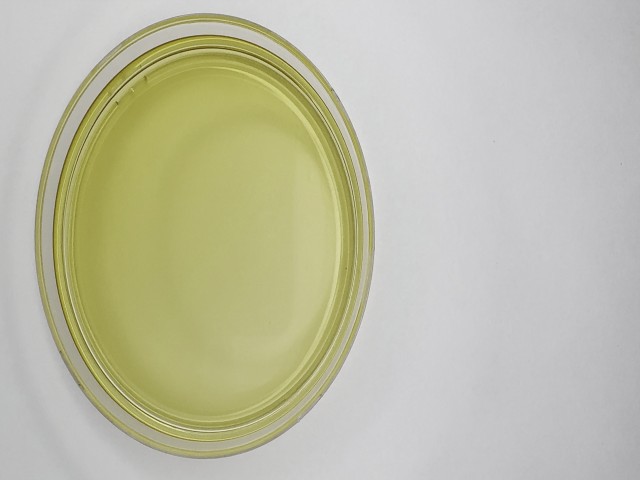} \\[-3pt]
& \multicolumn{2}{c}{%
  \colorbox{black!5}{\tiny
    PV\,=\,16.4 ~ p-AV\,=\,21.8 ~ Totox\,=\,54.5 ~ Temp\,=\,119\textdegree F}} \\
\end{tabular}
\vspace{-5pt}
\caption{Dataset samples. \textbf{Top:} fresh oil (good).
  \textbf{Bottom:} degraded oil (replace).
  Each frame carries paired thermal/RGB images with four regression targets.}
\label{fig:dataset_samples}
\vspace{-10pt}
\end{figure}

\noindent \textbf{Non-Destructive Inspection of Edible Oils.}
Non-destructive testing (NDT) for edible-oil quality aims to replace
wet-chemistry assays with rapid sensor-based alternatives.
NIR spectroscopy combined with PLS regression is the most established approach,
achieving $R^2>0.97$ for peroxide value prediction across multiple oil
classes~\cite{cayuela2025nir, ottaway2021spectroscopy, mehany2026rapidmonitoringand,
alalawi2021pv}, though reliable performance depends on careful
calibration and spectral pre-treatment.
Hyperspectral imaging extends spectral analysis to spatial domains for
adulteration detection~\cite{aqeel2025spectralbandselection}, but these
systems remain laboratory-bound~\cite{li2025recentadvancesand}.
Other sensing modalities (electronic noses~\cite{cozzolino2022enose,
sberveglieri2023review}, dielectric sensors~\cite{lizhi2019dielectric})
provide sample-level readings without spatial resolution.

Camera-based approaches offer a path toward spatially resolved
inspection.
RGB systems correlate CIE $L^*a^*b^*$ features with FFA/TPM
($R^2 > 0.91$) and achieve 95--97\% freshness
classification~\cite{udomkun2019cvs, naser2023rgb,
mancini2022olive}, and CNN-based classifiers further advance visual
food-quality evaluation~\cite{wang2025cnn, zhou2019deeplearning}.
Thermal imaging has broad utility in food quality
assessment~\cite{vadivambal2011thermal}, with recent work applying CNNs
to infrared thermograms for oil-type
identification~\cite{pirola2024infrared, kopelevich2024infrared}, yet
these approaches classify oil \emph{type} rather than regressing
chemical oxidation indices.

\noindent \textbf{Domain Adaptation and Shortcut Mitigation.}
Domain shift from heterogeneous acquisition pipelines is a recurring
challenge in applied vision.
Domain-Adversarial Neural Networks (DANN)~\cite{ganin2016dann} address
this through a Gradient Reversal Layer (GRL) that implements a minimax
objective: the domain classifier learns to predict domain identity while
the GRL reverses gradients during backpropagation, forcing the encoder
to produce domain-invariant representations.
Multi-discriminator variants improve robustness at the cost of training
complexity~\cite{gilani2024adversarialtrainingbased}, and recent work in
microscopy shows that adversarial alignment recovers target-domain
performance under optical and magnification
changes~\cite{bhattacharya2025enhancingaimicroscopy}.
Compared to MMD minimization, CORAL, and adversarial discriminators with
separate source/target networks, DANN requires no architectural
separation and trains end-to-end with the task loss.

In parallel, work on shortcut learning has shown that models frequently
exploit spurious acquisition-driven cues (camera fingerprints,
compression artifacts, frequency-domain
signatures)~\cite{sahidullah2025shortcutlearningin}, yielding strong
in-domain accuracy but brittle cross-domain generalization.
In our setting, we repurpose DANN not for cross-dataset transfer but for
intra-dataset regularization: each video/camera constitutes a domain,
and the GRL forces the encoder to discard camera-specific patterns.

\noindent \textbf{Self-Supervised and Masked Representation Learning.}
Self-supervised learning (SSL) has become a dominant strategy for
label-efficient representation learning.
Contrastive methods such as SimCLR~\cite{chen2020simclr} and
DINO~\cite{caron2021dino} learn transferable features by maximising
agreement between augmented views, while
DINOv2~\cite{oquab2023dinov2} scales this approach to produce
general-purpose visual features from large curated datasets.
Masked Autoencoders (MAE)~\cite{he2022mae} show that reconstructing
heavily masked (75\%) patches from a ViT
encoder~\cite{dosovitskiy2020vit} yields strong spatial
representations transferable with very few labels.
VideoMAE~\cite{tong2022videomae} extends MAE to video via temporal tube
masking, which we adapt for paired RGB-thermal cross-modal context in our
RGB-MAE Encoder.
These methods produce general features but are optimized for
natural-image semantics, providing no mechanism to suppress acquisition
shortcuts when labels correlate with camera identity.

\noindent \textbf{RGB-Thermal Fusion and Multi-Task Dense Prediction.}
RGB-thermal (RGB-T) fusion is widely studied for robustness under
challenging illumination.
Dense prediction systems typically use dual encoders with multi-level
fusion and a shared decoder.
CMX~\cite{zhang2023cmx} introduces cross-modal feature rectification
modules that exchange information between RGB and auxiliary encoders at
multiple scales, while CMNeXt~\cite{zhang2023cmnext} extends this with a
self-query hub layer that selects informative features across an
arbitrary number of modalities.
Both are designed for outdoor scene understanding (urban driving, indoor
scenes) and do not address camera-fingerprint confounds in industrial
inspection.
As a lighter alternative to cross-attention fusion, Feature-wise Linear Modulation
(FiLM)~\cite{perez2018film} offers efficient conditioning
via learned per-channel scale and shift at $O(NC)$ cost.

On the multi-task side, jointly training dense and scalar objectives
provides representation sharing and implicit
regularization~\cite{ruder2017overview, kendall2018mtl}, but
multi-task optimization can be dominated by task imbalance and
conflicting gradients~\cite{yu2020pcgrad, chen2018gradnorm}.
We sidestep this by gradient-isolating each task head and routing
detached post-fusion features to the regression branch.

FryNet bridges these gaps by unifying spatially resolved segmentation with chemical regression under domain-adversarial training that suppresses camera-specific shortcuts.

\section{Method}
\label{sec:method}

\subsection{Architecture Overview}
\label{sec:arch}

FryNet is a dual-stream, multi-task model that jointly segments
degraded oil regions, classifies oil serviceability, and regresses
chemical quality indicators from paired FLIR thermal and RGB inputs
(Fig.~\ref{fig:architecture}).
The model receives a FLIR thermal frame and a co-registered RGB frame, both resized to $512\!\times\!512$.
A thermal encoder extracts multi-scale spatial features from the FLIR
input; an RGB encoder learns chemically grounded representations via
auxiliary objectives; a lightweight fusion module conditions
thermal features on RGB context; and a multi-task decode head produces
segmentation maps and regression predictions.
Both encoders are regularized with gradient reversal~\cite{ganin2016dann} to suppress camera-fingerprint shortcuts (Sec.~\ref{sec:dann}).


\paragraph{ThermalMiT-B2}
The thermal stream uses MiT-B2~\cite{xie2021segformer}, whose
hierarchical overlapping-patch design yields a four-stage feature
pyramid $\{F_i\}_{i=1}^{4}$ at strides $\{4,8,16,32\}$ with
channel dimensions $[64, 128, 320, 512]$.

A lightweight thermal attention block is inserted after each stage.
Each block applies
(1)~\emph{Thermal Channel Attention} (TCA), a CBAM-style~\cite{woo2018cbam}
channel-recalibration module that upweights thermally informative
channels, and
(2)~\emph{Thermal Spatial Attention} (TSA), a $7\!\times\!7$
spatial gate that highlights high-gradient boundary regions
between fresh and oxidized oil.
Both initialize near-identity, adding $<$1\% parameters and
preserving pretrained representations.


\paragraph{RGB-MAE Encoder}
RGB captures complementary chemical signals invisible to thermal
imaging: Maillard browning, carotenoid degradation, and surface foam
texture are optical proxies for oxidation products
(PV, p-AV)~\cite{udomkun2019cvs}.
To extract these cues, we use an \emph{RGB-MAE Encoder}, a lightweight
6-layer Vision Transformer~\cite{dosovitskiy2021vit} with
embed\_dim\,=\,256 and patch size\,=\,16, trained from scratch,
that processes a paired RGB frame and produces a dense feature map
$\mathbf{s}_{\text{ctx}}\!\in\!\mathbb{R}^{256\times 32\times 32}$.

The encoder is trained jointly with the rest of the model via two
auxiliary objectives that reuse existing annotations:

\emph{(i) Masked autoencoding} ($\mathcal{L}_{\text{MAE}}$).
We adapt VideoMAE~\cite{tong2022videomae} to our paired setting:
75\% of the current FLIR frame's patch tokens are randomly masked,
and the encoder receives the concatenation of the remaining visible
FLIR tokens and all RGB context tokens; both modalities pass through
the same shared transformer blocks.
A lightweight two-layer pixel decoder (256$\to$128$\to$patch\_dim)
then reconstructs the masked FLIR patches via L1 loss against the
original pixels.
Because visible FLIR tokens alone cannot reconstruct the masked
patches, the encoder must leverage RGB context, learning cross-modal
correspondences without explicit supervision.

\emph{(ii) Chemical alignment} ($\mathcal{L}_{\text{chem}}$).
After encoding, we global-average-pool the context-frame token
sequence to obtain a single descriptor, then project it through a
two-layer MLP (256$\to$128$\to$3) onto z-score predictions of
[p-AV, Totox, temperature].
A Huber loss against the ground-truth chemical measurements grounds
the RGB encoder in oxidation chemistry.
This head is auxiliary: it is active only during training and adds no
cost at inference.


\paragraph{Multi-scale feature fusion}
The four backbone feature maps $\{F_i\}$ span a 8$\times$ resolution
range (stride 4 to stride 32).
Following SegFormer~\cite{xie2021segformer}, we unify them into a single
representation: each $F_i$ is projected to $C\!=\!256$ channels via a
$1\!\times\!1$ convolution with batch normalization, upsampled to the
finest resolution ($H/4\!\times\!W/4$) via bilinear interpolation, and
concatenated along the channel axis.
A final $1\!\times\!1$ convolution reduces the $4C$-channel tensor back
to $C$ channels, yielding
$F_{\text{ms}}\!\in\!\mathbb{R}^{C\times H/4\times W/4}$.

\paragraph{FiLM fusion}
$F_{\text{ms}}$ encodes spatial temperature structure but lacks
chemical context; $\mathbf{s}_{\text{ctx}}$ encodes oxidation chemistry
but at coarse resolution ($H/16$).
We bridge them via Feature-wise Linear Modulation
(FiLM)~\cite{perez2018film}, chosen over cross-attention for its
5$\times$ fewer parameters (50\,k vs.\ 263\,k at $C\!=\!256$).
Because the two feature maps differ in spatial resolution,
$F_{\text{ms}}$ is first downsampled to match
$\mathbf{s}_{\text{ctx}}$ ($H/16\!\times\!W/16$), fused, then
upsampled back to $H/4\!\times\!W/4$.
Fusion proceeds in three steps:
\begin{align}
  (\boldsymbol{\gamma},\boldsymbol{\beta})
    &= \mathrm{MLP}\!\bigl(\mathrm{GAP}(\mathbf{s}_{\text{ctx}})\bigr)
       \in \mathbb{R}^{C}, \label{eq:film_gb}\\
  \mathbf{m}
    &= F_{\text{ms}}\,(1+\boldsymbol{\gamma}) + \boldsymbol{\beta},
       \label{eq:film_mod}\\
  g &= \sigma\!\bigl(W_g\,\mathbf{s}_{\text{ctx}}\bigr)
       \in [0,1]^{H'\times W'}, \label{eq:film_gate}\\
  \mathbf{F}_{\text{fused}}
    &= \mathrm{GN}\!\bigl(\alpha\,\mathbf{m}\odot g
       + (1{-}\alpha)\,F_{\text{ms}}\bigr),
       \label{eq:film_blend}
\end{align}
where $\boldsymbol{\gamma},\boldsymbol{\beta}$ are per-channel
scale and shift derived from the global RGB descriptor
(Eq.~\ref{eq:film_gb}), $g$ is a learned spatial gate that selectively
emphasizes regions where RGB indicates oxidation activity
(Eq.~\ref{eq:film_gate}), and $\alpha$ is a learnable blend scalar
(Eq.~\ref{eq:film_blend}).
We initialize $\boldsymbol{\gamma}\!=\!0$, $\boldsymbol{\beta}\!=\!0$, and gate bias\,=\,4.0 ($\sigma(4)\!\approx\!0.98$) so the module begins as a near-identity pass-through, preventing the early training collapse we observed with unconstrained cross-attention fusion.

\paragraph{Multi-task decode head}
The fused representation $\mathbf{F}_{\text{fused}}$ feeds three
task-specific branches:

\emph{Segmentation.}
A $1\!\times\!1$ convolution produces per-pixel logits
$\hat{S}\!\in\!\mathbb{R}^{3\times H\times W}$
(\emph{background}\,/\,\emph{good}\,/\,\emph{replace}), upsampled to input resolution at
loss time.
An auxiliary head ($3\!\times\!3$ conv + $1\!\times\!1$ conv on $F_1$)
provides a secondary segmentation signal.
Frame-level classification is derived from the segmentation map by majority vote over the oil region. Since each frame contains a single class, no separate classification head is needed.

\emph{Regression.}
Four two-layer MLPs ($256{\to}256{\to}1$)
predict z-scored values of PV, p-AV, Totox, and temperature from
global-average-pooled \emph{fused} features $\mathbf{F}_{\text{fused}}$.
Regression heads consume fused features,
giving them access to RGB chemical context
(Sec.~\ref{sec:ablation}).
Gradient isolation is maintained via stop-gradient: regression
loss trains only the MLP heads, not the backbone or fusion module,
preventing regression gradients from corrupting segmentation
features.

\subsection{Domain Adaptation via Gradient Reversal}
\label{sec:dann}

\paragraph{The camera-fingerprint problem}
As described in Sec.~\ref{sec:intro}, the camera-fingerprint shortcut
lets models memorize sensor identity instead of oxidation chemistry.
An analogous shortcut exists in the RGB stream, where per-device
white balance, lens distortion, and sensor noise provide
camera-identifying features.
Suppressing both shortcuts is necessary not only for segmentation
generalization but also for accurate chemical regression:
without adversarial regularization, backbone features remain
video-dominant even when segmentation performance is high,
leaving regression heads unable to recover accurate predictions
(Sec.~\ref{sec:ablation}).

\paragraph{Gradient reversal}
We adopt Domain-Adversarial Neural Networks
(DANN)~\cite{ganin2016dann}, which enforce feature invariance to
domain identity via a minimax objective:
\begin{equation}
  \min_\theta \max_d \;\mathcal{L}_\text{task}(\theta)
    - \lambda\,\mathcal{L}_\text{dann}(d,\theta),
\end{equation}
where $\theta$ are encoder parameters and $d$ are domain-classifier
parameters.
A Gradient Reversal Layer (GRL) multiplies the domain-classifier
gradient by $-\alpha$ during backpropagation, driving the encoder to
produce features that maximally confuse video-ID prediction while
minimizing task loss.

\paragraph{Dual-stream design}
Because the FLIR backbone and RGB encoder carry independent
camera fingerprints, we apply separate DANN heads to each:
\begin{align}
  \mathcal{L}_\text{dann}
    &= \text{CE}\!\bigl(\text{MLP}(\text{GRL}(\text{GAP}(F_4))),\;
       v_\text{id}\bigr), \\
  \mathcal{L}_\text{rgb-dann}
    &= \text{CE}\!\bigl(\text{MLP}(\text{GRL}(\text{GAP}(
       \mathbf{s}_\text{ctx}))),\; v_\text{id}\bigr),
\end{align}
where $v_\text{id} \in \{0,\ldots,19\}$ indexes the 20 training
videos (Sec.~\ref{sec:dataset}), and each MLP is a two-layer classifier
(Linear-BN-ReLU-Dropout($p{=}0.5$)-Linear).
Both heads share $\lambda\!=\!0.1$ and $\alpha\!=\!1.0$.
Validation and test videos never contribute to either DANN loss.
The effect of each component is analyzed in Sec.~\ref{sec:ablation}.

\paragraph{Total training objective}
The full loss combines segmentation, regression, self-supervised, and domain-adaptation terms:
\begin{equation}
  \mathcal{L} = \mathcal{L}_{\text{seg}} + \mathcal{L}_{\text{aux}}
    + \mathcal{L}_{\text{reg}}
    + \mathcal{L}_{\text{MAE}} + \mathcal{L}_{\text{chem}}
    + \mathcal{L}_{\text{dann}} + \mathcal{L}_{\text{rgb-dann}},
\end{equation}
where $\mathcal{L}_{\text{reg}} = \mathcal{L}_{\text{PV}} + \mathcal{L}_{\text{p-AV}} + \mathcal{L}_{\text{Totox}} + \mathcal{L}_{\text{temp}}$.
All regression and chemical alignment losses use Huber loss; segmentation uses cross-entropy.
Loss weights: segmentation 0.1, auxiliary 0.4, Totox 1.0, PV/p-AV/temperature 0.5 each, MAE and chemical alignment 0.3 each, both DANN heads 0.1.


\section{Dataset}
\label{sec:dataset}

\paragraph{Experimental design}
The dataset originates from controlled deep-frying experiments designed
to monitor lipid oxidation in edible oils during thermal processing.
Two oil types commonly used for deep-frying in the United States were
selected: corn oil (9~frying cycles) and canola oil
(5~frying cycles), yielding 14~cycles total.
Each frying cycle lasted approximately 10~minutes.
Each cycle is captured twice, once with fresh oil (\emph{before})
and once after prolonged frying (\emph{after}), producing 28~video
sequences.

\begin{table}[t]
\centering
\caption{
  \textbf{FryNet dataset statistics.}
  All splits are strictly video-disjoint.
  Good\,/\,replace distribution: 3{,}909\,/\,3{,}317 frames.
}
\label{tab:dataset}
\small
\setlength{\tabcolsep}{4pt}
\begin{tabular}{lrrrrrr}
\toprule
Oil Type & Videos & Frames & \% & Train & Val & Test \\
\midrule
Corn oil   & 18 & 4{,}861 & 67.3 & 3{,}408 & 660 & 793 \\
Canola oil & 10 & 2{,}365 & 32.7 & 1{,}912 & 241 & 212 \\
\midrule
\rowcolor{rowgroup}
\textbf{Total} & \textbf{28} & \textbf{7{,}226} & \textbf{100} & \textbf{5{,}320} & \textbf{901} & \textbf{1{,}005} \\
\bottomrule
\end{tabular}
\vspace{-10pt}
\end{table}

\paragraph{Frying procedure}
Falafel samples were fried in a laboratory deep fryer containing 1.5\,L
of oil maintained at $180 \pm 2\,^\circ$C, monitored by a thermocouple
probe.
Repeated batch frying over several hours induced progressive thermal
degradation and oxidation.
Oil samples were collected before and after each frying session for
chemical analysis.

\begin{figure}[t]
\centering
\includegraphics[width=\linewidth]{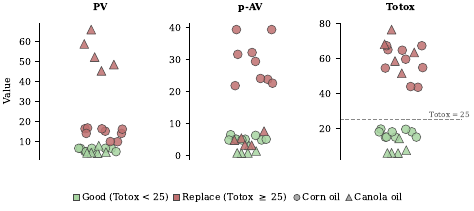}
\caption{
  \textbf{Distribution of chemical oxidation indices} across 28 oil
  samples.  Colors indicate segmentation class (green = good,
  Totox\,$<$\,25; red = replace, Totox\,$\geq$\,25); markers
  distinguish oil type ($\circ$ = corn, $\triangle$ = canola).
  Dashed line marks the Totox\,=\,25 classification threshold.
}
\label{fig:chemical_scatter}
\vspace{-10pt}
\end{figure}

\paragraph{Chemical evaluation}
Three oxidation indices were measured for each oil sample:
\textbf{(i)}~Peroxide Value (PV), an indicator of primary oxidation
products, determined by iodometric titration following AOAC
protocols~\cite{aydin2021antioxidant};
\textbf{(ii)}~p-Anisidine Value (p-AV), measuring secondary oxidation
products (aldehydes), determined by spectrophotometry following AOCS
Cd~18-90~\cite{aocs2017pav}; and
\textbf{(iii)}~Total Oxidation Value (Totox), combining both indicators
as $\text{Totox} = 2 \times \text{PV} + \text{p-AV}$~\cite{shahidi2002methods}.
Frying temperature ($^\circ$F) is extracted per-frame via OCR from the
FLIR on-screen display.
Across the 28~videos, Totox ranges from 5.8 (fresh canola) to 76.6
(degraded canola), spanning an order-of-magnitude variation in
oxidation load.
Figure~\ref{fig:chemical_scatter} shows the distribution of all three
indices across the 28~oil samples, colored by segmentation class.
The two classes are well separated in Totox space, while PV and p-AV
individually reveal distinct oxidation profiles between corn and canola
oils: canola shows higher primary oxidation (PV) whereas corn shows
elevated secondary products (p-AV).

\paragraph{Imaging}
For each oil sample, an iPhone~15~Pro records a 4K RGB video of the
oil surface at room temperature, then the oil is heated and a
FLIR~GF77 long-wave infrared (LWIR) camera captures the thermal
video at $640\!\times\!480$ resolution (Figure~\ref{fig:dataset_samples}).
Both modalities capture the same oil batch and chemical state.
Frames are extracted at 1\,fps and resized to $640\!\times\!480$; RGB
portrait frames are rotated to landscape and temporally aligned to their
FLIR counterparts via uniform index resampling across each video pair.

\paragraph{Annotation}
Each of the 7{,}226 frames carries four annotation types
(Table~\ref{tab:dataset}):
\textbf{(i)}~a pixel-level mask delineating the oil surface from
background, generated by Segment Anything~2 (SAM2,
Hiera-Large)~\cite{ravi2024sam2} with bounding-box prompts obtained from
Otsu thresholding of the thermal image.
FLIR on-screen display elements (battery icon, temperature readout,
color-scale bar) are excluded from masks by cropping 9\% top, 8\%
bottom, and 9\% right margins, so OSD regions are labeled as
background in the segmentation ground truth;
\textbf{(ii)}~a ternary segmentation label (\emph{background},
\emph{good}: Totox\,$<$\,25, \emph{replace}: Totox\,$\geq$\,25),
applied uniformly to all oil-surface pixels within a frame;
\textbf{(iii)}~four continuous regression targets (PV, p-AV, Totox, and
temperature), z-score normalized using training-set statistics; and
\textbf{(iv)}~a video-sequence identifier linking each frame to its
camera and frying batch.

\begin{table*}[t]
\centering
\caption{Main results on the 1{,}005-frame test set.
         $^\ddagger$Classification derived from segmentation
         majority vote (not a separately learned head).
         All regression MAEs are in raw (denormalized) units:
         PV (meq~O$_2$/kg), p-AV, Totox, and temperature ($^\circ$F).
         \textbf{Bold} = best.}
\label{tab:main}
\small
\setlength{\tabcolsep}{4pt}
\begin{tabular}{ll rrr ccc ccccc}
\toprule
& &
  \multicolumn{3}{c}{\cellcolor{rowgroup}\textbf{Efficiency}}
& \multicolumn{3}{c}{\cellcolor{colseg}\textbf{Seg / Cls}}
& \multicolumn{5}{c}{\cellcolor{colreg}\textbf{Regression MAE $\downarrow$}} \\
\cmidrule(lr){3-5} \cmidrule(lr){6-8} \cmidrule(lr){9-13}
Method & Backbone
  & \cellcolor{rowgroup!50}Params & \cellcolor{rowgroup!50}GFLOPs & \cellcolor{rowgroup!50}FPS
  & \cellcolor{colseg!50}mIoU & \cellcolor{colseg!50}mF1 & \cellcolor{colseg!50}Cls$^\ddagger$
  & \cellcolor{colreg!50}PV & \cellcolor{colreg!50}p-AV & \cellcolor{colreg!50}Totox & \cellcolor{colreg!50}Temp & \cellcolor{colreg!50}Mean \\
\midrule
\rowcolor{rowgroup}
\multicolumn{13}{l}{\textit{Single-modal baselines (thermal only)}} \\
SegFormer       & MiT-B2      & 24.9M & 25.3 & 66.2 & 61.80 & 72.58 & 64.7 & 4.74 &  6.34 & 14.35 & 4.10 &  7.38 \\
ConvNeXt-B      & ConvNeXt-B  & 88.5M & 85.8 & 59.0 & 62.36 & 73.40 & 63.0 & 4.64 &  8.88 & 14.64 & 4.98 &  8.29 \\
DeepLabV3       & ResNet-50   & 24.9M & 110.0 & 130.0 & 59.76 & 70.78 & 67.1 & 3.71 & 13.58 & 14.26 & 2.46 &  8.50 \\
Swin-S          & Swin-S      & 49.6M & 54.5 & 40.4 & 58.82 & 69.78 & 61.2 & 3.87 & 10.43 & 16.64 & 3.34 &  8.57 \\
DINOv2          & ViT-B       & 87.8M & 119.0 & 46.4 & 54.15 & 64.56 & 80.1 & 3.24 & 11.88 & 11.99 & 7.58 &  8.67 \\
\midrule
\rowcolor{rowgroup}
\multicolumn{13}{l}{\textit{Multi-modal fusion baselines (RGB + thermal)}} \\
CMX             & MiT-B2$\times$2 & 66.7M & 78.3 & 31.1 & 18.12 & 27.19 & 47.8 & 6.41 & 35.93 & 23.25 & 16.92 & 20.62 \\
CMNeXt          & MiT-B2$\times$2 & 58.8M & 74.2 & 31.0 & 76.39 & 85.62 & 78.9 & 4.18 & 10.72 & 10.43 &  6.76 &  8.02 \\
\rowcolor{rowours}
FryNet (Ours)  & MiT-B2      & \textbf{31.0M} & \textbf{30.3} & 47.1 & \textbf{98.97} & \textbf{99.48} & \textbf{100} & \textbf{2.66} & \textbf{1.98} & \textbf{2.86} & \textbf{1.80} & \textbf{2.32} \\
\bottomrule
\end{tabular}
\vspace{-10pt}
\end{table*}

\paragraph{Splits and camera-fingerprint constraint}
The dataset is partitioned into strictly video-disjoint subsets
(Table~\ref{tab:dataset}): no video appears in more than one split,
ensuring that evaluation measures generalization to \emph{entirely unseen
oil batches and cameras}.
An important consequence of per-video labeling is that every video is single-class.
A model that memorizes per-camera sensor signatures (fixed-pattern noise,
vignetting gradients, thermal bias) can therefore achieve near-perfect
training accuracy without learning any oxidation chemistry.
We analyze this camera-fingerprint shortcut and our adversarial fix in
Sec.~\ref{sec:dann}.

\section{Experiments}
\label{sec:experiments}

\subsection{Implementation Details}

We used the mmsegmentation~\cite{mmseg2020} codebase and trained on
a single NVIDIA A100-40\,GB GPU.
All backbones are initialized from ImageNet-1K pre-trained weights,
while decode heads, regression heads, and the RGB-MAE Encoder are
randomly initialized.
During training, we applied random resize (0.5--2.0$\times$),
horizontal flipping, photometric distortion, and random cropping to
$512{\times}512$.
We trained all models using AdamW~\cite{loshchilov2019adamw}
(lr $6{\times}10^{-5}$, weight decay $10^{-2}$) for 40{,}000
iterations with a batch size of 4, using a linear warm-up over
1{,}500 iterations followed by polynomial decay (power~1.0).
All single-modal baselines and FryNet variants share identical
training hyperparameters.
CMX and CMNeXt retain their published optimizer and learning-rate
schedule (AdamW, lr $6{\times}10^{-5}$, PolyLR with power 0.9,
30 epochs) but are adapted to our dataset at native FLIR resolution
($480{\times}640$) with added regression heads for fair multi-task
comparison.
We report segmentation performance using mIoU and mF1, classification
using majority-vote accuracy from the segmentation map, and regression
using per-target MAE in raw denormalized units.

\paragraph{Baselines}
The single-modal thermal baselines are
SegFormer~\cite{xie2021segformer} (MiT-B2),
ConvNeXt-B~\cite{liu2022convnet},
DeepLabV3~\cite{chen2017deeplabv3} (ResNet-50),
Swin-S~\cite{liu2021swin}, and
DINOv2-ViT-B~\cite{oquab2024dinov2}.
For multi-modal comparison we evaluate
CMX~\cite{zhang2023cmx} and CMNeXt~\cite{zhang2023cmnext}
(Sec.~\ref{sec:related}), both using dual MiT-B2 encoders with
the same paired RGB+thermal inputs as FryNet; neither includes
domain adaptation or auxiliary learning, testing whether
multi-modal fusion alone overcomes the camera-fingerprint
shortcut.

\subsection{Quantitative Results}

We report segmentation, classification, and regression performance
on the held-out 1{,}005-frame test set (4~videos)
in Table~\ref{tab:main}.

\begin{figure*}[t]
  \centering
  \setlength{\tabcolsep}{1pt}
  \scriptsize
  \begin{tabular}{@{}c cccccccccc@{}}
  & Image & GT & SegFormer & DeepLabV3 & ConvNeXt & Swin & DINOv2 & CMX & CMNeXt & \textbf{FryNet} \\[2pt]
  \raisebox{12pt}{\smash{\rotatebox[origin=c]{90}{\scriptsize Good}}} &
  \includegraphics[width=\segw]{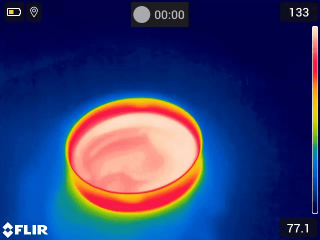} &
  \includegraphics[width=\segw]{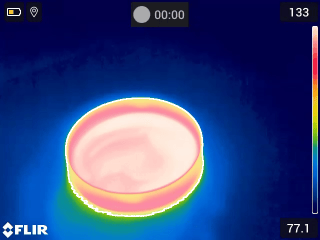} &
  \includegraphics[width=\segw]{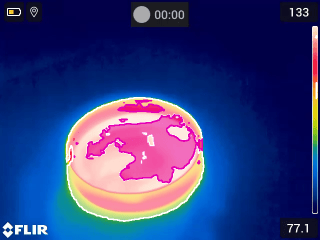} &
  \includegraphics[width=\segw]{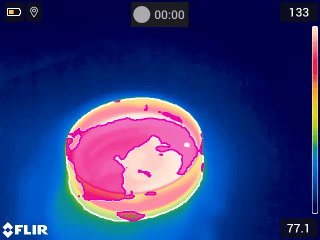} &
  \includegraphics[width=\segw]{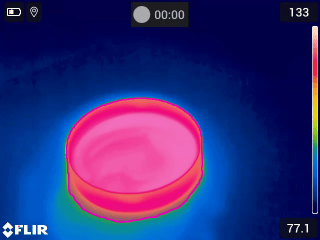} &
  \includegraphics[width=\segw]{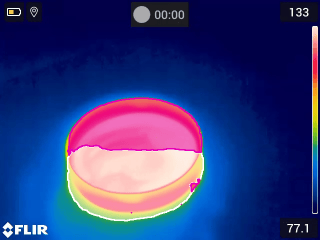} &
  \includegraphics[width=\segw]{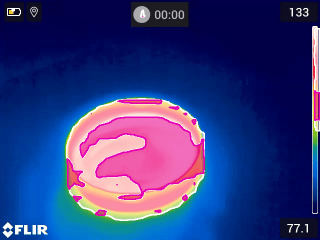} &
  \includegraphics[width=\segw]{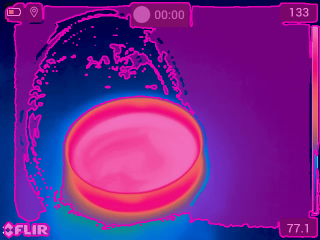} &
  \includegraphics[width=\segw]{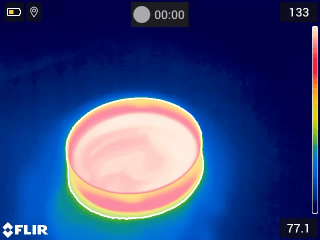} &
  \includegraphics[width=\segw]{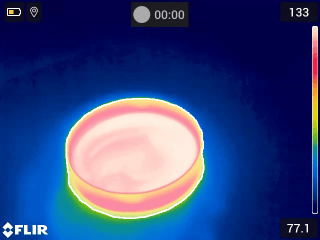} \\[-1pt]
  & & & \cmark & \cmark & \xmark & \cmark & \cmark & \xmark & \cmark & \cmark \\[1pt]
  \raisebox{12pt}{\smash{\rotatebox[origin=c]{90}{\scriptsize Good}}} &
  \includegraphics[width=\segw]{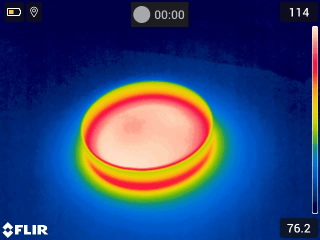} &
  \includegraphics[width=\segw]{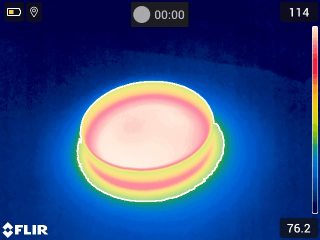} &
  \includegraphics[width=\segw]{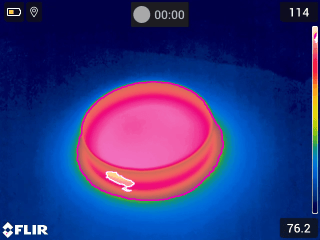} &
  \includegraphics[width=\segw]{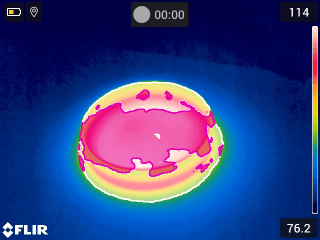} &
  \includegraphics[width=\segw]{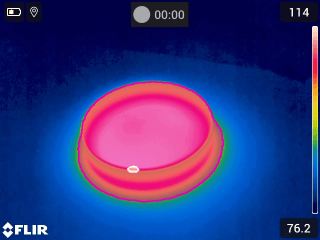} &
  \includegraphics[width=\segw]{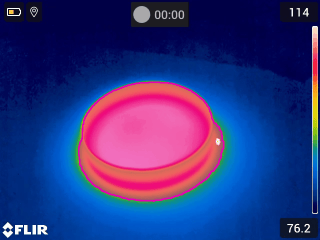} &
  \includegraphics[width=\segw]{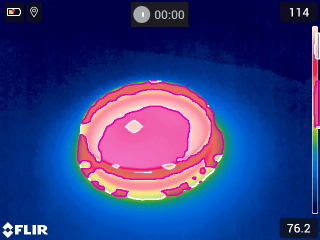} &
  \includegraphics[width=\segw]{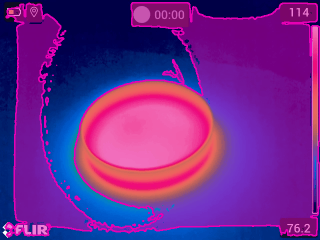} &
  \includegraphics[width=\segw]{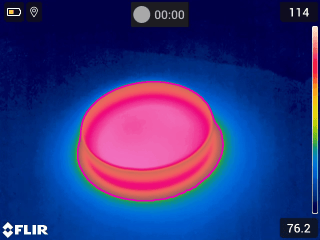} &
  \includegraphics[width=\segw]{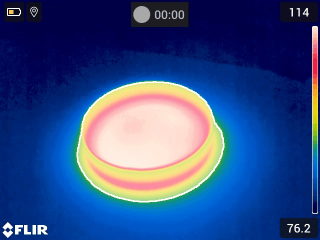} \\[-1pt]
  & & & \xmark & \xmark & \xmark & \xmark & \xmark & \xmark & \xmark & \cmark \\[1pt]
  \raisebox{12pt}{\smash{\rotatebox[origin=c]{90}{\scriptsize Replace}}} &
  \includegraphics[width=\segw]{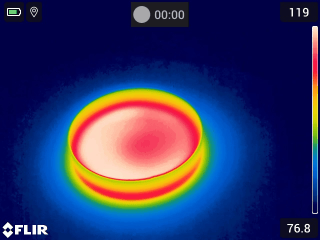} &
  \includegraphics[width=\segw]{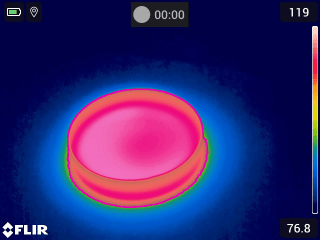} &
  \includegraphics[width=\segw]{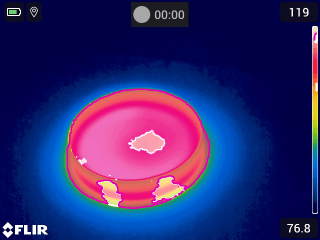} &
  \includegraphics[width=\segw]{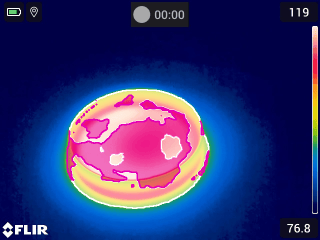} &
  \includegraphics[width=\segw]{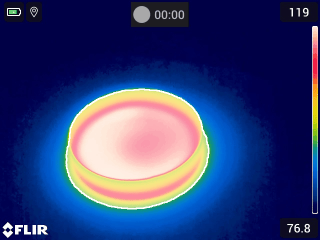} &
  \includegraphics[width=\segw]{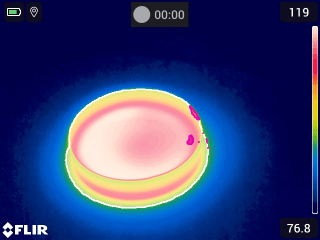} &
  \includegraphics[width=\segw]{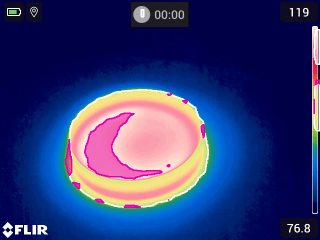} &
  \includegraphics[width=\segw]{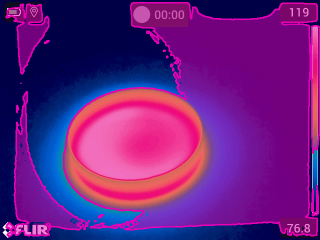} &
  \includegraphics[width=\segw]{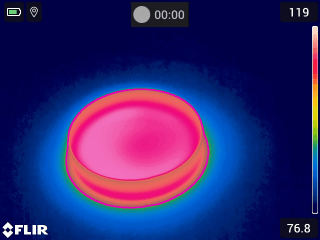} &
  \includegraphics[width=\segw]{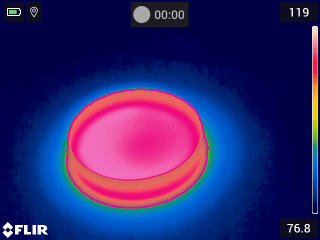} \\[-1pt]
  & & & \cmark & \xmark & \xmark & \xmark & \xmark & \cmark & \cmark & \cmark \\[1pt]
  \raisebox{12pt}{\smash{\rotatebox[origin=c]{90}{\scriptsize Replace}}} &
  \includegraphics[width=\segw]{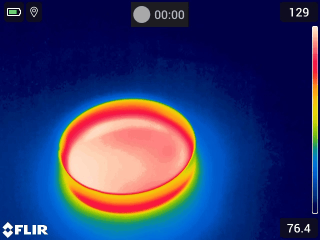} &
  \includegraphics[width=\segw]{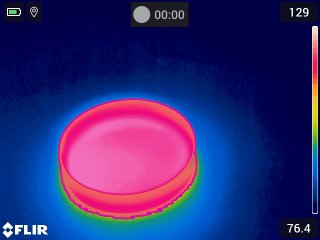} &
  \includegraphics[width=\segw]{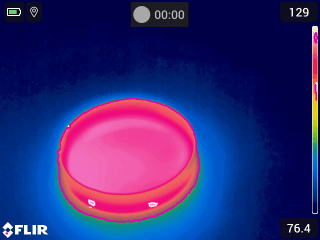} &
  \includegraphics[width=\segw]{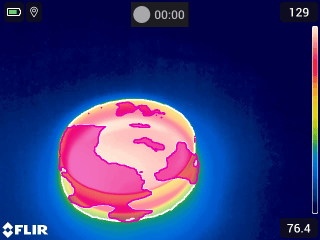} &
  \includegraphics[width=\segw]{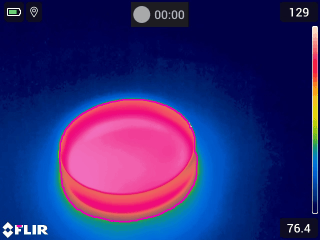} &
  \includegraphics[width=\segw]{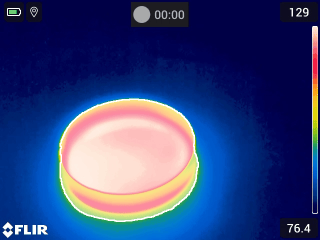} &
  \includegraphics[width=\segw]{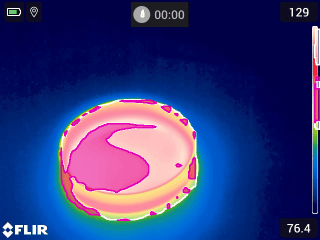} &
  \includegraphics[width=\segw]{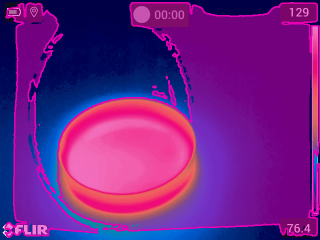} &
  \includegraphics[width=\segw]{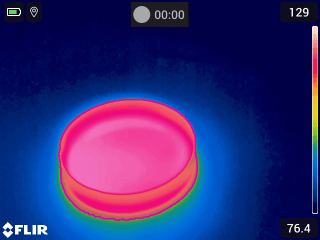} &
  \includegraphics[width=\segw]{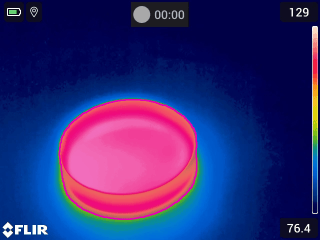} \\[-1pt]
  & & & \cmark & \xmark & \cmark & \xmark & \xmark & \cmark & \cmark & \cmark \\
  \end{tabular}
  \vspace{-5pt}
  \caption{Qualitative segmentation comparison on representative test frames
  from two \emph{good}-class and two \emph{replace}-class videos.
  Predictions are overlaid on the thermal image:
  \textbf{white} = good oil;
  \textcolor[RGB]{230,0,180}{\textbf{magenta}} = replace oil.
  \cmark/\xmark{} indicates whether the dominant predicted oil class
  matches ground truth.}
  \label{fig:seg_comparison}
  \vspace{-10pt}
\end{figure*}

\paragraph{Single-modal baselines}
Table~\ref{tab:main} summarizes segmentation, classification, and
regression results along with efficiency metrics for all methods.
In the top block, all five single-modal thermal baselines plateau
below 63\% mIoU, consistent with the camera-fingerprint bottleneck
identified in Sec.~\ref{sec:dann}.
SegFormer yields the best single-modal mean MAE of 7.38, while
Totox MAE remains consistently above~11 across all baselines.

\begin{figure}[b]
\centering
\includegraphics[width=\columnwidth]{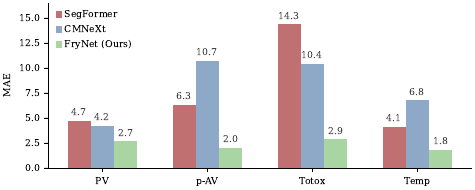}
\caption{Per-target regression MAE on the test set.}
\label{fig:reg_bars}
\vspace{-10pt}
\end{figure}

\paragraph{Multi-modal fusion baselines}
In the middle block, we evaluate two multi-modal RGB-thermal baselines
using dual MiT-B2 encoders with the same paired inputs as FryNet.
CMX collapses to 18.12\% mIoU, worse than majority-class prediction,
as fusion without adversarial suppression amplifies rather than
mitigates the camera-fingerprint shortcut.
CMNeXt performs better at 76.39\% mIoU owing to its FRM module,
but still falls 22.6~pp below FryNet while using 2$\times$ more
parameters and GFLOPs.

\paragraph{FryNet}
As shown in the bottom block of Table~\ref{tab:main}, FryNet achieves
98.97\% mIoU and 100\% classification accuracy using only 31.0M
parameters and 30.3 GFLOPs, outperforming all baselines by a wide
margin at lower computational cost than all comparable multi-task
methods (Figure~\ref{fig:miou_efficiency}).
Mean regression MAE of 2.32 represents a 3.2$\times$ improvement
over the best single-modal baseline SegFormer at 7.38 and a
3.5$\times$ improvement over CMNeXt at 8.02.
As shown in Figure~\ref{fig:reg_bars}, FryNet achieves the lowest
MAE on all four regression targets, with Totox MAE of 2.86 compared
to 10.43 for CMNeXt and 14.35 for SegFormer.

\paragraph{Qualitative segmentation results}
Figure~\ref{fig:seg_comparison} shows representative predictions
from all eight methods on four test videos.
Single-modal baselines show \emph{class inversion}: the predicted
class flips between videos depending on camera identity, with no two
baselines agreeing on the same frame.
DeepLabV3 and DINOv2 fragment the oil surface into patches of both
classes, following camera vignetting gradients rather than any physical
boundary.
CMX predicts \emph{replace} over both oil and background across all
videos, while CMNeXt fails on one of four videos (row~2).
FryNet assigns $>$99\% of oil pixels to the correct class on all
four videos with spatially uniform masks.

Table~\ref{tab:ablation} isolates the contribution of each FryNet
component through incremental construction, controlled removal,
DA method and fusion method comparisons, and architectural alternatives.
All ablations use the SegFormer (MiT-B2) backbone.

\begin{figure}[b]
\centering
\includegraphics[width=\columnwidth]{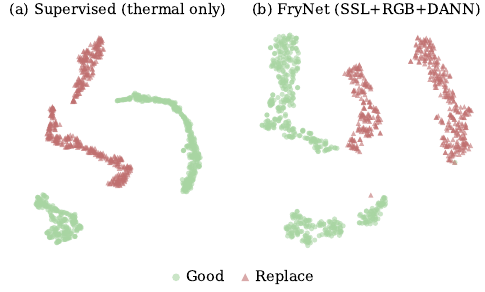}
\vspace{-5pt}
\caption{t-SNE of backbone features on the test set, colored by
         class (\textcolor[RGB]{168,213,162}{good} vs.\
         \textcolor[RGB]{192,112,112}{replace}).}
\label{fig:tsne}
\vspace{-10pt}
\end{figure}

\begin{table*}[t]
\centering
\caption{Ablation study (SegFormer MiT-B2, test set).
         Enc = RGB-MAE Encoder active.
         Th/R = Thermal/RGB DANN active.
         $\checkmark^*$ = modified encoder variant.
         Fused = post-fusion features routed to regression heads.
         \colorbox{rowours}{Shaded} row = FryNet complete pipeline.}
\label{tab:ablation}
\small
\setlength{\tabcolsep}{4pt}
\begin{tabular}{l c cc c c ccc ccccc}
\toprule
& & \multicolumn{2}{c}{\cellcolor{coldann}\textbf{DANN}} & & &
  \multicolumn{3}{c}{\cellcolor{colseg}\textbf{Seg / Cls}}
& \multicolumn{5}{c}{\cellcolor{colreg}\textbf{Regression MAE $\downarrow$}} \\
\cmidrule(lr){3-4} \cmidrule(lr){7-9} \cmidrule(lr){10-14}
Variant & Enc & \cellcolor{coldann!50}Th & \cellcolor{coldann!50}R & RGB & Fused
  & \cellcolor{colseg!50}mIoU & \cellcolor{colseg!50}mF1 & \cellcolor{colseg!50}Cls
  & \cellcolor{colreg!50}PV & \cellcolor{colreg!50}p-AV & \cellcolor{colreg!50}Totox & \cellcolor{colreg!50}Temp & \cellcolor{colreg!50}Mean \\
\midrule
\rowcolor{rowgroup}
\multicolumn{14}{l}{\textit{Pipeline construction}} \\
Thermal only
  & ---   & ---          & ---          & ---   & ---
  & 61.80 & 72.58 & 64.7
  & 4.74 & 6.34 & 14.35 & 4.10 & 7.38 \\
+ DANN
  & ---   & $\checkmark$ & ---          & ---   & ---
  & 52.59 & 61.44 & 51.4
  & 4.49 & 10.75 & 18.41 & 3.92 & 9.39 \\
+ RGB encoder + dual-DANN
  & $\checkmark$ & $\checkmark$ & $\checkmark$ & $\checkmark$ & ---
  & 99.23 & 99.61 & 100
  & 5.29 & 7.78 & 18.20 & 2.41 & 8.42 \\
\rowcolor{rowours}
+ Fused regression
  & $\checkmark$ & $\checkmark$ & $\checkmark$ & $\checkmark$ & $\checkmark$
  & 98.97 & 99.48 & 100
  & 2.66 & 1.98 & 2.86 & 1.80 & 2.32 \\
\midrule
\rowcolor{rowgroup}
\multicolumn{14}{l}{\textit{Component removals}} \\
$-$ Thermal DANN
  & $\checkmark$ & ---          & $\checkmark$ & $\checkmark$ & $\checkmark$
  & 99.05 & 99.52 & 100
  & 0.85 & 2.26 & 2.37 & 4.15 & 2.41 \\
$-$ Chem alignment
  & $\checkmark$ & $\checkmark$ & $\checkmark$ & $\checkmark$ & $\checkmark$
  & 64.17 & 75.18 & 63.9
  & 4.76 & 5.56 & 13.85 & 3.33 & 6.88 \\
$-$ All DANN
  & $\checkmark$ & ---          & ---          & $\checkmark$ & $\checkmark$
  & 98.45 & 99.21 & 100
  & 3.20 & 2.83 & 5.98 & 14.98 & 6.75 \\
\midrule
\rowcolor{rowgroup}
\multicolumn{14}{l}{\textit{DA method comparison}} \\
\rowcolor{rowours}
GRL
  & $\checkmark$ & $\checkmark$ & $\checkmark$ & $\checkmark$ & $\checkmark$
  & 98.97 & 99.48 & 100
  & 2.66 & 1.98 & 2.86 & 1.80 & 2.32 \\
MMD
  & $\checkmark$ & $\checkmark$ & $\checkmark$ & $\checkmark$ & $\checkmark$
  & 99.25 & 99.62 & 100
  & 1.77 & 2.50 & 2.56 & 7.99 & 3.71 \\
CORAL
  & $\checkmark$ & $\checkmark$ & $\checkmark$ & $\checkmark$ & $\checkmark$
  & 99.29 & 99.65 & 100
  & 1.53 & 2.51 & 4.42 & 5.55 & 3.50 \\
\midrule
\rowcolor{rowgroup}
\multicolumn{14}{l}{\textit{Fusion method comparison (no fused reg)}} \\
FiLM
  & $\checkmark$ & $\checkmark$ & $\checkmark$ & $\checkmark$ & ---
  & 99.23 & 99.61 & 100
  & 5.29 & 7.78 & 18.20 & 2.41 & 8.42 \\
Attention
  & $\checkmark$ & $\checkmark$ & $\checkmark$ & $\checkmark$ & ---
  & 98.97 & 99.48 & 100
  & 5.04 & 8.66 & 16.27 & 5.93 & 8.98 \\
Concat
  & $\checkmark$ & $\checkmark$ & $\checkmark$ & $\checkmark$ & ---
  & 99.00 & 99.50 & 100
  & 5.19 & 6.22 & 16.29 & 6.45 & 8.54 \\
\midrule
\rowcolor{rowgroup}
\multicolumn{14}{l}{\textit{Architecture alternatives}} \\
Dual MiT-B2 (early fusion)
  & ---   & $\checkmark$ & ---          & $\checkmark$ & ---
  & 98.81 & 99.40 & 100
  & 0.42 & 10.58 & 10.55 & 1.45 & 5.75 \\
TemporalMeanEncoder
  & $\checkmark^*$ & $\checkmark$ & $\checkmark$ & $\checkmark$ & ---
  & 98.44 & 99.21 & 99.8
  & 5.34 & 7.61 & 17.02 & 6.89 & 9.21 \\
No MAE loss
  & $\checkmark^*$ & $\checkmark$ & $\checkmark$ & $\checkmark$ & ---
  & 98.99 & 99.49 & 100
  & 5.11 & 6.81 & 14.37 & 9.42 & 8.92 \\
\bottomrule
\end{tabular}
\vspace{-10pt}
\end{table*}

\subsection{Ablation Study}
\label{sec:ablation}

\paragraph{Pipeline construction}
In the top part of Table~\ref{tab:ablation}, we report the incremental
construction of FryNet.
The thermal-only baseline yields 61.80\% mIoU and 7.38 mean MAE.
Adding DANN alone worsens segmentation to 52.59\% mIoU,
indicating that adversarial training is counterproductive without
cross-modal context to compensate.
Adding the RGB-MAE Encoder with dual-DANN recovers segmentation
to 99.23\% mIoU with perfect classification, but regression remains
poor at 8.42 mean MAE without access to fused features.
Enabling fused regression drops mean MAE to 2.32,
a 3.6$\times$ improvement, while maintaining 98.97\% mIoU.

\paragraph{Component removals}
As shown in the second block, removing the thermal DANN while
keeping the RGB DANN has negligible impact, raising mean MAE only
from 2.32 to 2.41.
In contrast, removing chemical alignment collapses segmentation
to 64.17\% mIoU despite both DANN branches remaining active,
indicating that it provides a critical early learning signal that
guides the RGB encoder before DANN stabilizes.
Removing all DANN preserves 98.45\% mIoU but degrades regression
to 6.75 mean MAE, with temperature MAE rising from 1.80 to 14.98,
consistent with the feature-space analysis in Figure~\ref{fig:tsne}.

\paragraph{DA method comparison}
In the third block, we compare GRL against MMD and CORAL applied
to both streams.
MMD and CORAL yield slightly higher mIoU of 99.25\% and 99.29\%,
but substantially worse mean MAE of 3.71 and 3.50 vs.\ 2.32 for GRL,
driven by elevated temperature MAE of 7.99 and 5.55 vs.\ 1.80.
GRL provides the best overall regression, making it the default choice.

\paragraph{Fusion method comparison}
As reported in the fourth block, we compare three fusion strategies
under identical settings with dual-DANN and no fused regression.
FiLM achieves the lowest mean MAE of 8.42 vs.\ 8.54 for concat
and 8.98 for attention, the highest mIoU of 99.23\%, and is the
most parameter-efficient with 5$\times$ fewer parameters than
cross-attention (Sec.~\ref{sec:arch}).

\paragraph{Architecture alternatives}
Finally, in the bottom part of Table~\ref{tab:ablation}, a
dual-backbone early-fusion model using two parallel MiT-B2 encoders
achieves 98.81\% mIoU and 5.75 mean MAE, which is 2.5$\times$
worse than FryNet's 2.32, indicating that doubling backbone capacity
cannot substitute for learned cross-modal representations.
Replacing the RGB-MAE Encoder with temporal mean pooling degrades
mean MAE to 9.21, confirming that learned representations are critical.
Disabling MAE reconstruction loss yields 8.92 mean MAE, showing it
provides complementary signal beyond chemical alignment alone.

\subsection{Limitations}

The test set comprises four video sequences (1{,}005 frames)
from four unseen oil batches.
Because every video is single-class, a single misclassified
video shifts mIoU by ${\sim}$25~pp; results should be interpreted
at the video level.
Classification is derived from segmentation majority vote
(Table~\ref{tab:main}, $^\ddagger$), so it does not test
independent classification capability.
Dataset scale (28~videos from one facility) may not capture
the full diversity of frying conditions and fryer geometries.
Future work will expand the dataset to cover more diverse frying conditions, oil types, and fryer geometries beyond the single laboratory setup used in this study.

\section{Conclusion}
\label{sec:conclusion}

We presented FryNet, a multi-modal RGB-thermal framework for
non-destructive frying oil quality inspection that addresses the
camera-fingerprint shortcut, a failure mode in which models
memorize sensor-specific signatures rather than oxidation chemistry.
Our Dual-Encoder DANN suppresses this shortcut in both thermal and RGB
streams via adversarial video-identity regularization, while a
chemistry-grounded alignment loss stabilizes training and a fused
regression routing strategy enables a mean regression MAE of 2.32,
a 3.2$\times$ improvement over the best single-modal baseline.
FryNet achieves 98.97\% mIoU with
100\% classification accuracy, outperforming five single-modal and two
multi-modal baselines by wide margins.

{
    \small
    \bibliographystyle{ieeenat_fullname}
    \bibliography{main}
}


\end{document}